\documentclass[conference]{IEEEtran}
\IEEEoverridecommandlockouts
\usepackage{cite}
\usepackage{amsmath,amssymb,amsfonts}
\usepackage{graphicx}
\usepackage{textcomp}
\usepackage{xcolor}
\usepackage{float}
\usepackage{url}
\usepackage[caption=false,font=footnotesize]{subfig}

\usepackage{cuted}    
\usepackage{capt-of}  
\usepackage{url}  
\usepackage{hyperref}
\usepackage{subcaption}

\def\BibTeX{{\rm B\kern-.05em{\sc i\kern-.025em b}\kern-.08em    \kern-.1667em\lower.7ex\hbox{E}\kern-.125emX}}

\begin{document}
\title{An Explainable FFT-Based Spatial-Frequency Fusion Framework for Deepfake Detection}



\author{ \IEEEauthorblockN{Pamela Kirui, Cho Hyuk, Qingzhong Liu, and Haodi Jiang\textsuperscript{*}} \IEEEauthorblockA{\textit{Department of Computer Science} 
\\ \textit{Sam Houston State University}
\\ Huntsville, Texas, USA\\ haodi.jiang@shsu.edu} 
\thanks{\textsuperscript{*}Corresponding author: Haodi Jiang.} 
}

\maketitle

\begin{abstract}
Deepfake generation has raised growing concerns regarding digital media authenticity, misinformation, identity fraud, and public trust. Recent studies show that combining spatial and frequency features leads to stronger detection results than using either independently. 
This paper presents MSCA-FFT, a Fast Fourier Transform (FFT)-based multi-scale cross-attention framework for image-level deepfake detection. The model combines a partially fine-tuned Xception spatial branch with 
an FFT-based frequency branch. The frequency branch processes the log-scaled FFT magnitude spectrum through shallow convolutional layers, avoiding inverse frequency-to-image reconstruction used in DCT-based pipelines. The spatial and frequency representations are refined by transformer encoders, fused through cross-attention, and passed to an MLP classifier for real/fake prediction.
Experimental results show that MSCA-FFT 
achieves consistently higher performance than
the DCT-based state-of-the-art spatial-frequency fusion method and the compared baseline models. The ablation study further indicates that the FFT-based frequency branch provides complementary spectral cues when fused with spatial features. In addition, FFT-based frequency analysis and Grad-CAM/LIME explanations show consistent evidence around manipulation-sensitive facial regions, including the eyes, mouth, nose, and facial boundaries.
\end{abstract}

\begin{IEEEkeywords}
Deepfake face detection, Deep learning, Fast Fourier Transform (FFT), Spatial--frequency fusion
\end{IEEEkeywords}

\section{Introduction}\label{sec:introduction}

Deepfake media generated by Generative Adversarial Networks (GANs), diffusion models, and other advanced synthesis techniques has become increasingly difficult to distinguish from authentic content. These methods can generate highly realistic facial images and videos, reducing the visibility of conventional manipulation artifacts and making visual inspection unreliable. As a result, deepfake media has raised significant concerns in disinformation, political manipulation, identity fraud, and 
cybercrime~\cite{Masood2023,VaccariChadwick2020}. 
Reliable deepfake detection remains challenging because manipulation artifacts are often subtle, compression can obscure discriminative cues, and generation methods vary substantially across datasets and manipulation pipelines.

Deepfake detection methods have evolved from early hand-crafted artifact analysis to deep learning-based approaches that exploit spatial, frequency, and spatial-frequency cues. Early methods examined visible irregularities such as inconsistent head pose and abnormal blinking, but these cues are often sensitive to pose variation, compression, and facial expression changes~\cite{Yang2019HeadPose}. With the development of deep learning, CNN-based models became dominant in spatial-domain detection. MesoNet~\cite{afchar2018mesonet} emphasizes computational efficiency, while Xception~\cite{chollet2017xception} has become a strong baseline for capturing fine-grained texture artifacts and broader contextual patterns. However, spatial-domain methods may overlook subtle or repeated manipulation traces that are more distinguishable in transformed representations~\cite{durall2020watch}.

Frequency-domain analysis provides complementary evidence by examining spectral patterns that may be less visible in the pixel domain. Prior studies have shown that manipulated content can exhibit abnormal frequency distributions or periodic artifacts~\cite{yu2019attributing,durall2020watch}. F3-Net further demonstrated that local frequency-aware features can improve forged face detection by capturing manipulation artifacts beyond RGB appearance~\cite{qian2020thinking}. Other studies have incorporated wavelet-based representations to improve robustness against compression and texture bias~\cite{guo2023constructing}. Nevertheless, frequency-only methods can also be affected by lighting variation, textured regions, background patterns, and compression artifacts, which may introduce spectral changes unrelated to manipulation.

Because spatial and frequency cues provide complementary information, recent deepfake detectors have increasingly combined both types of features to improve detection robustness~\cite{qiu2025d2fusion,uddin2025msca}. Existing multi-branch methods often align spatial and frequency representations using attention-based fusion or cross-attention mechanisms~\cite{chen2021crossvit,xu2025few}. Although such designs can improve detection performance, they may also increase model complexity and computational cost. In particular, the recent DCT-based method~\cite{uddin2025msca} reconstructs selected frequency information into an image-like representation using inverse DCT before fusion with the spatial branch, introducing additional processing steps. In contrast, the proposed method extracts frequency features directly from FFT representations, avoiding inverse reconstruction and simplifying the frequency-processing pipeline.

Several benchmark datasets have been introduced to support deepfake detection research, including FaceForensics++~\cite{rossler2019faceforensicspp}, the DeepFake Detection Challenge dataset~\cite{dolhansky2019deepfake}, and Celeb-DF~\cite{li2020celeb}. 
This study focuses on Celeb-DF and FaceForensics++, which provide complementary evaluation settings: Celeb-DF contains high-quality and natural-looking manipulations, while FaceForensics++ provides an additional benchmark with multiple manipulation methods and different dataset characteristics.

Using these benchmarks, this study develops an image-level deepfake detection pipeline for real/fake classification. 
For Celeb-DF (v2), a preprocessed face-image dataset is used, while for FaceForensics++ face crops are extracted from video frames using Multi-task Cascaded Convolutional Networks (MTCNN)~\cite{zhang2016mtcnn}. 
The resulting face images are resized, normalized, organized into real and fake classes, and evaluated using a consistent five-fold protocol. 
The proposed model, referred to as MSCA-FFT, is an FFT-based multi-scale cross-attention framework that combines a pre-trained Xception backbone for spatial feature extraction~\cite{chollet2017xception} with a direct FFT-based frequency branch for spectral artifact detection~\cite{tan2024frequency,frank2020leveraging}. 
The resulting spatial and frequency representations are refined by Transformer encoders and integrated through a cross-attention fusion module followed by an MLP classification head for binary real/fake prediction.

The main contributions of this paper are summarized as follows:
\begin{itemize}
\item A direct FFT-based frequency pathway is introduced to derive features from FFT magnitude representations through shallow convolutional layers, avoiding inverse frequency-to-image reconstruction before spatial-frequency fusion.
\item An FFT-based multi-scale cross-attention framework, termed MSCA-FFT, is developed by combining a pre-trained Xception spatial branch, an FFT-based frequency branch, Transformer-based feature modeling, and cross-attention fusion for image-level real/fake classification.
\item An explainability analysis is provided using Grad-CAM and LIME to examine the model's decision behavior, showing that the learned evidence is concentrated around manipulation-sensitive regions such as the eyes, nose, mouth, and facial boundaries.
\end{itemize}

The rest of this paper is organized as follows: Section~\ref{sec:method} describes the proposed methodology. Section~\ref{sec:Experiment and Result Analysis} presents the experimental results and ablation studies. Section~\ref{sec:conclusion} concludes the paper.

\section{Methodology}\label{sec:method}

\begin{figure*}[ht!]
  \centering
  \includegraphics[width=0.99\textwidth]{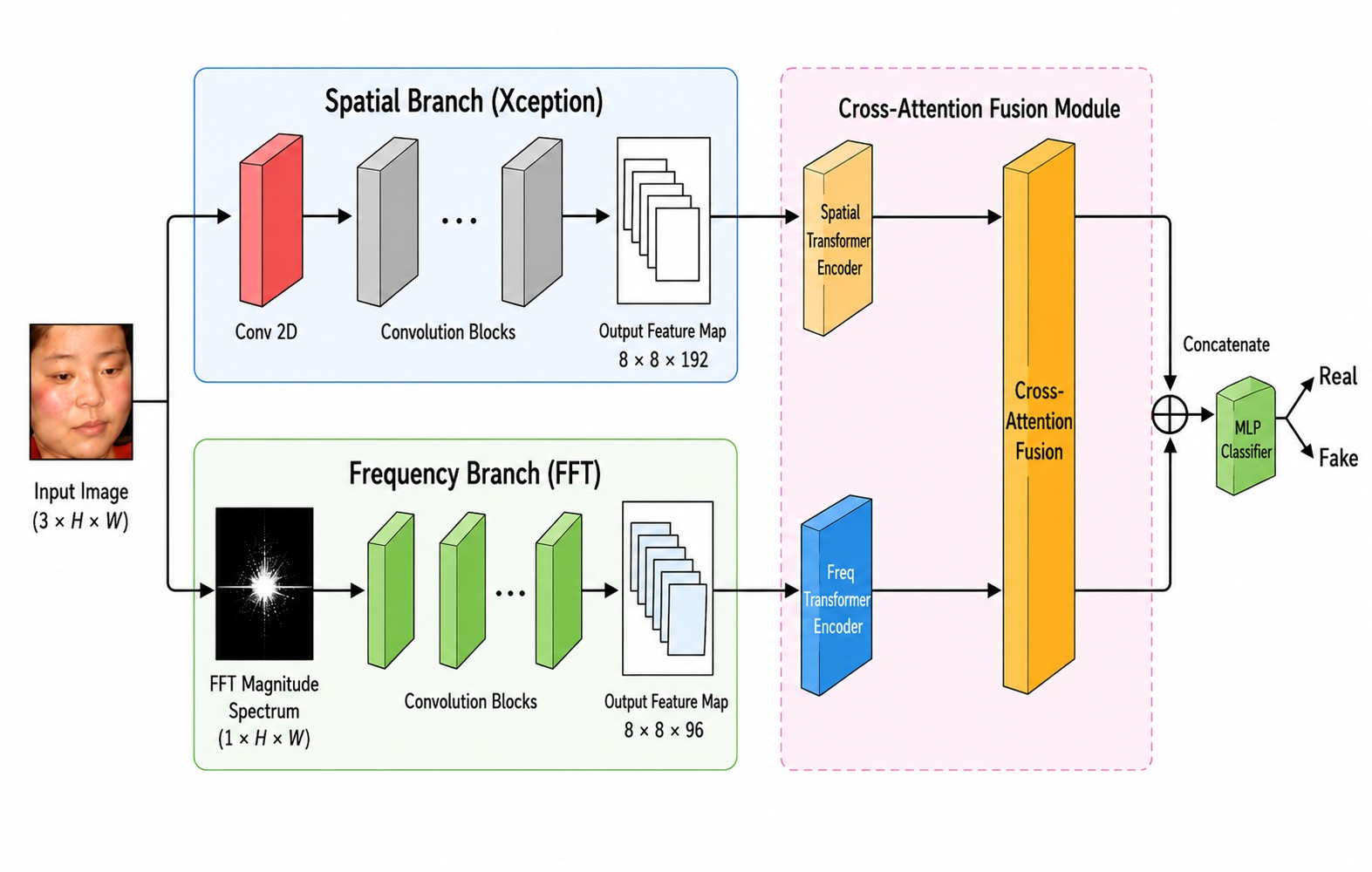}
  \vspace{-5mm}
  \caption{Architecture of the proposed spatial-frequency deepfake face detection framework. The input face image is processed by an Xception-based spatial branch and an FFT-based frequency branch. The spatial branch extracts image-domain texture and structural features, while the frequency branch learns spectral artifacts from the FFT magnitude spectrum. The resulting embeddings are refined by transformer encoders, integrated through a cross-attention fusion module, concatenated, and passed to a classifier for real/fake prediction.}
  \label{fig:workflow}
\end{figure*}

\subsection{Overall Framework}
As shown in Fig.~\ref{fig:workflow}, the proposed framework consists of three main components: an Xception-based spatial feature extraction branch, an FFT-based frequency feature extraction branch, and a cross-attention fusion module followed by a multilayer perceptron (MLP) classifier. Given an input face image, the spatial branch captures local visual cues such as boundary distortions, texture irregularities, shading inconsistencies, and facial blending artifacts. In parallel, the frequency branch transforms the input image into an FFT magnitude representation and learns spectral patterns that may be less visible in the pixel domain~\cite{qian2020thinking,durall2020watch}. The two feature streams are integrated through cross-attention to form a joint spatial-frequency representation, which is then passed to an MLP classification head for binary real/fake prediction.

\subsection{Spatial Feature Extraction Module}

Spatial features are important for identifying local texture, boundary, shading, and structural inconsistencies that distinguish authentic faces from manipulated ones~\cite{Li2019Warping,Li2020FaceXray,Yang2019HeadPose}. To extract these cues, the proposed spatial branch uses a pre-trained Xception backbone~\cite{chollet2017xception}, which employs depthwise separable convolutions to perform channel-wise spatial filtering followed by pointwise cross-channel integration. This design enables efficient extraction of fine-grained spatial patterns with fewer parameters than standard convolutions. In the proposed implementation, 80\% of the Xception layers are frozen, and the top 20\% are fine-tuned to adapt the backbone to the deepfake detection task while reducing overfitting risk. 
Given an input face image, the spatial branch produces an output feature map of size $8 \times 8 \times 192$, which is then passed to the spatial transformer encoder before being integrated in the cross-attention fusion module.

\subsection{FFT-Based Frequency Feature Extraction Module}

The frequency branch is designed to capture spectral artifacts that are often difficult to observe directly in the spatial domain~\cite{durall2020watch,frank2020leveraging}. 
A recent DCT-based method transforms the image into the frequency domain, applies multi-scale frequency filters, and reconstructs image-like representations through inverse transforms before neural feature extraction~\cite{uddin2025msca}. 
In contrast, the proposed frequency branch avoids this reconstruction step and directly processes the FFT magnitude spectrum using shallow convolutional layers.

Given an input face image, the image representation is transformed using the two-dimensional Fast Fourier Transform (FFT).  The magnitude spectrum is then computed and log-scaled to reduce the dynamic range:

\begin{equation}
F = \mathcal{F}(I),
\end{equation}

\begin{equation}
M = \log \left(1 + |F|\right),
\end{equation}

where $I$ denotes the input face image representation, $\mathcal{F}(\cdot)$ represents the 2D FFT operation, and $M$ denotes the log-scaled FFT magnitude spectrum. 
This FFT representation is passed directly to the CNN-based frequency branch without applying an inverse FFT.

Each convolution block in the frequency branch consists of three convolutional layers that process the log-scaled FFT magnitude representation. 
Within each block, the first convolutional layer applies 64 filters with same padding, while the second and third convolutional layers apply 128 and 96 filters, respectively, with stride-2 downsampling to gradually reduce the spatial resolution. 
Each convolutional layer is followed by batch normalization and ReLU activation.
The output is resized to a fixed spatial resolution of $8 \times 8$, producing an $8 \times 8 \times 96$ frequency feature map. 
This feature map serves as the frequency-domain representation for the subsequent feature modeling and cross-attention fusion stages.

To interpret the frequency-domain representation used by the proposed branch, the FFT magnitude spectrum is examined across low-, mid-, and high-frequency regions.
Recent frequency-masking studies for deepfake detection show that different frequency regions contain useful cues for identifying manipulated 
content~\cite{doloriel2024frequency,shi2025customized}.
Low-frequency components mainly reflect global illumination, intensity distribution, and coarse structural information. Mid-frequency components capture texture patterns and local structural inconsistencies, while high-frequency components emphasize fine details such as edges, compression traces, and sensor-like noise patterns.

\begin{figure*}[ht]
    \centering
    \includegraphics[width=0.95\linewidth]{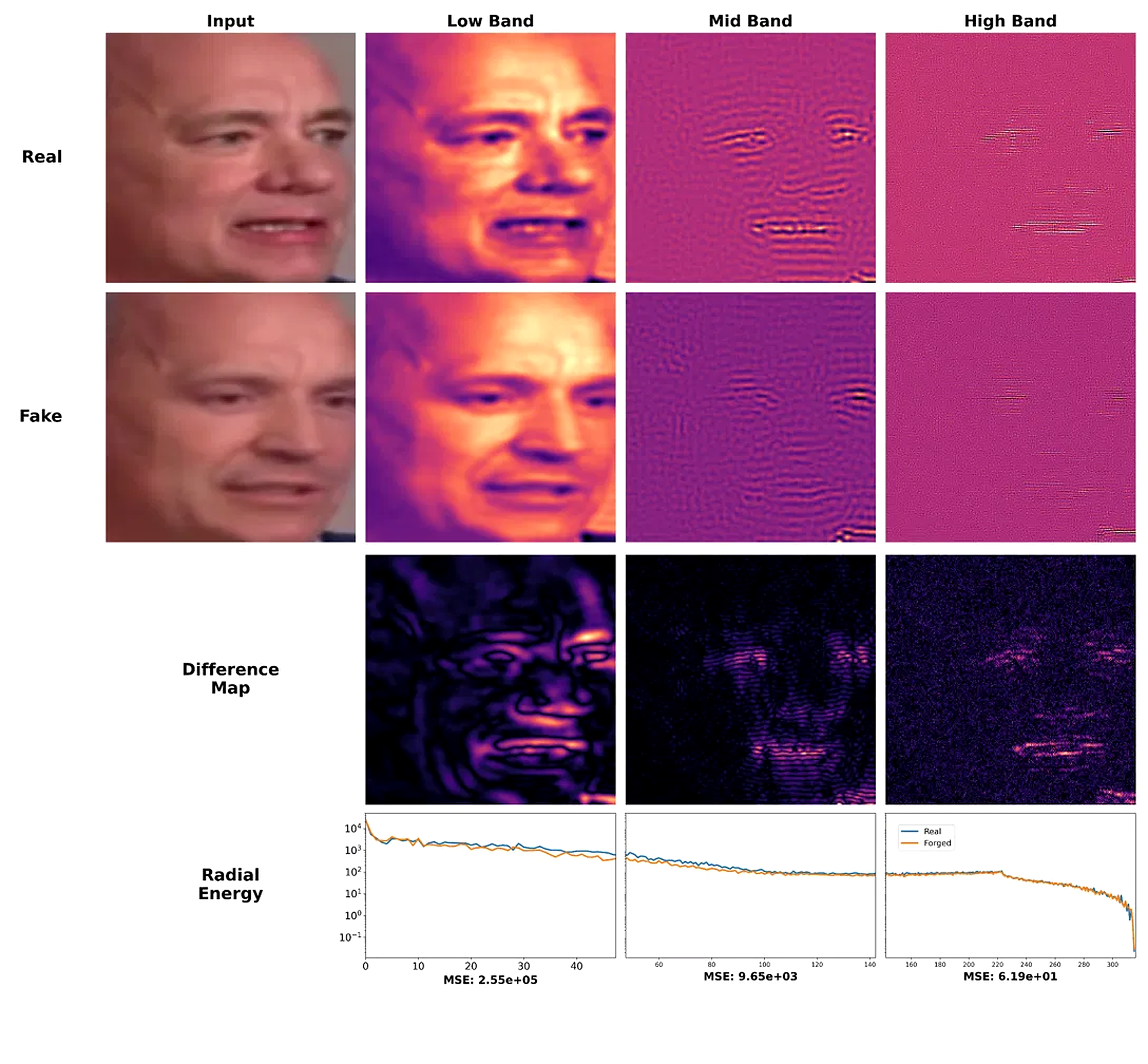}
    \vspace{-5mm}
    \caption{FFT-based frequency analysis of authentic and forged facial images across low-, mid-, and high-frequency regions. The first two rows show the band-specific frequency responses for real and fake samples, while the third row presents the corresponding difference maps to highlight spectral discrepancies. The final row shows radial energy distributions on a logarithmic scale for each frequency band, illustrating differences in global intensity structure, local texture patterns, and fine-scale details.The reported MSE values correspond to the single real–fake image pair shown.}
    \label{fig:frequency}
\end{figure*}

As illustrated in Fig.~\ref{fig:frequency}, the band-specific frequency responses and difference maps reveal visible spectral differences between authentic and forged samples. The first two rows present the low-, mid-, and high-frequency responses, while the third row shows the corresponding difference maps, highlighting manipulation-related discrepancies around facial regions such as the eyes and mouth. The final row presents radial energy distributions on a logarithmic scale for each frequency band. The spectral differences are further quantified using Mean Squared Error (MSE). In the illustrated example, the low-, mid-, and high-frequency bands show MSE values of 254,954.11, 9,649.49, and 61.90, respectively. These results indicate that the FFT-based representation captures differences across global intensity structure, local texture patterns, and fine-scale details.

\subsection{Spatial-Frequency Fusion and Classification}

After spatial and frequency feature extraction, the two feature maps are processed by separate transformer encoders before fusion. 
The downstream transformer-based feature modeling, cross-attention fusion, and MLP classification structure follow a design similar to the DCT-based MSCA framework~\cite{uddin2025msca}. 
This setting allows the effect of replacing the DCT-based frequency representation with the proposed FFT-based frequency representation to be evaluated more directly.

The spatial feature map is passed to the spatial transformer encoder, which further models relationships among spatial features extracted from facial regions. 
Similarly, the frequency feature map is passed to the frequency transformer encoder, which captures relationships among spectral features from the FFT magnitude representation. 
This design allows each branch to refine its domain-specific representation before spatial-frequency fusion.
The encoded spatial and frequency representations are then passed to the cross-attention fusion module. 
Rather than relying only on direct concatenation, cross-attention allows the two branches to exchange complementary information. 
In this process, the spatial representation can incorporate frequency-aware spectral cues, while the frequency representation can be guided by spatially meaningful facial context.

The attended spatial and frequency representations are concatenated into a joint embedding and passed to an MLP classification head. 
The classifier consists of Layer Normalization, a fully connected layer with 256 hidden units, GELU activation, dropout, and a final two-unit output layer. 
The final real/fake prediction is obtained using softmax normalization.

\subsection{Training Configuration}  

The training setup includes AdamW optimizer with a learning rate of $5 \times 10^{-5}$, weight decay of $5 \times 10^{-5}$, and gradient clipping at norm 1.0. Categorical cross-entropy with label smoothing of 0.03 was used as the loss function. Training used a batch size of 32 for up to 50 epochs per fold under a 5-fold cross-validation. Early stopping monitored validation AUC with a patience of 10 epochs, and a learning 
rate scheduler reduced the learning rate by a factor of 0.5 after 5 epochs of no improvement, with a minimum of $1 \times 10^{-7}$.
Data augmentation included random horizontal flips, brightness variation of $\pm 10\%$, 
contrast and saturation scaling in $[0.9, 1.1]$, and random JPEG compression simulation.

\section{Experiments and Results}\label{sec:Experiment and Result Analysis}

\subsection{Datasets}

Two benchmark datasets were used in this study: Celeb-DF (v2) and FaceForensics++. 
Celeb-DF (v2) was used as the primary benchmark for model training and evaluation, whereas FaceForensics++ was used to further assess robustness across a different deepfake dataset. 
For both datasets, all compared methods followed the same five-fold cross-validation protocol with identical training settings. 
In each fold, one fold was held out for testing, while the remaining folds were used for model training and validation. 
All reported metrics were computed separately for each fold and then averaged across the five folds.

\subsubsection{Celeb-DF (V2)}
Celeb-DF (v2)~\cite{li2020celeb} is a widely used benchmark 
comprising 6,229 videos (590 real, 5,639 fake) across diverse 
subjects and generation methods.
The preprocessed image-level 
version~\cite{celebdfv2kaggle} was used, with faces detected 
via dlib and enhanced using Efficient Sub-Pixel Convolutional Neural Network (ESPCN) super-resolution, yielding 
101,491 face images (50,740 real, 50,751 fake).
All images were resized to $256\times256$.

\subsubsection{FaceForensics++}

FaceForensics++~\cite{rossler2019faceforensicspp} is a widely adopted benchmark for face manipulation detection. 
It contains 1,000 pristine YouTube videos and four types of manipulated videos generated using Deepfakes, Face2Face, FaceSwap, and NeuralTextures. 
In this study, FaceForensics++ was used as an additional benchmark to evaluate the robustness of the proposed model across different manipulation methods and dataset characteristics. 
Each video was sampled every 15 frames using OpenCV, and corrupted or near-duplicate frames were discarded. 
MTCNN was then used to detect and align faces, with bounding boxes slightly expanded to retain peripheral facial regions. 
A similarity transform was applied to correct in-plane rotation and normalize scale based on eye landmarks. 
The resulting face crops were resized to $256\times256$ and normalized to $[0,1]$.

\subsection{Evaluation Metrics}

We adopt five metrics to evaluate the performance of the deepfake detection models: accuracy, precision, recall, F1-score, and the area under the receiver operating characteristic curve (AUC). These metrics are computed using true positives (TP), true negatives (TN), false positives (FP), and false negatives (FN). In this binary real/fake classification task, fake samples are treated as the positive class. TP refers to fake samples that are correctly classified as fake, while TN refers to real samples that are correctly classified as real. FP refers to real samples that are incorrectly classified as fake, and FN refers to fake samples that are incorrectly classified as real.
AUC is included as a threshold-independent metric to evaluate the model's ability to distinguish between real and fake samples across different decision thresholds.

\textbf{Accuracy (ACC):} Accuracy measures the proportion of correctly classified samples among all testing samples. It provides an overall indicator of classification performance.

\begin{equation}
\text{ACC} = \frac{TP + TN}{TP + TN + FP + FN}
\label{eq:accuracy}
\end{equation}

\textbf{Precision (PREC):} Precision measures the proportion of samples predicted as fake that are actually fake. A higher precision indicates that the model produces fewer false alarms when identifying fake samples.

\begin{equation}
\text{PREC} = \frac{TP}{TP + FP}
\label{eq:precision}
\end{equation}

\textbf{Recall (REC):} Recall measures the proportion of actual fake samples that are correctly detected by the model. A higher recall indicates that the model is more effective at identifying manipulated samples.

\begin{equation}
\text{REC} = \frac{TP}{TP + FN}
\label{eq:recall}
\end{equation}

\textbf{F1-Score:} The F1-score is the harmonic mean of precision and recall. It provides a balanced measure of detection performance, especially when both false positives and false negatives are important.

\begin{equation}
\text{F1-Score} = \frac{2 \cdot TP}{2 \cdot TP + FP + FN}
\label{eq:f1_tp}
\end{equation}

\textbf{Area Under the ROC Curve (AUC):} The receiver operating characteristic (ROC) curve characterizes the relationship between the true positive rate (TPR) and false positive rate (FPR) across different classification thresholds. AUC measures the area under this curve and evaluates the model's overall ability to separate real and fake samples.

\begin{equation}
\text{TPR} = \frac{TP}{TP + FN}
\label{eq:tpr}
\end{equation}

\begin{equation}
\text{FPR} = \frac{FP}{FP + TN}
\label{eq:fpr}
\end{equation}

\begin{equation}
\text{AUC} = \int_0^1 \text{TPR}(\text{FPR}) \, d\text{FPR}
\label{eq:auc}
\end{equation}

\subsection{Performance Analysis}

Tables~\ref{tab:method_comparison}--\ref{tab:ffpp_results} summarize the quantitative performance of the proposed model. 
All numerical results reported in this and following sections were obtained using five-fold cross-validation, and the mean values across the five folds are reported.
In this study, MSCA denotes multi-scale cross-attention. MSCA-FFT refers to the proposed FFT-based variant, where FFT magnitude-spectrum features are used as the frequency-domain representation and fused with Xception-based spatial features through the MSCA structure. MSCA-DCT denotes the DCT-based counterpart proposed in~\cite{uddin2025msca}, which uses DCT-based frequency features within a similar MSCA-Xception framework.
For the baseline CNN models used in the comparison, pretrained weights were adopted, and all layers were fine-tuned; no layers were frozen.

\begin{table}[!h]
\centering
\caption{Comparison with existing and baseline methods on the Celeb-DF (v2) dataset.}
\label{tab:method_comparison}
\footnotesize
\begin{tabular*}{0.98\columnwidth}{@{\extracolsep{\fill}}lccccc}
\hline
\textbf{Method} & \textbf{AUC} & \textbf{Acc.} & \textbf{Prec.} & \textbf{Rec.} & \textbf{F1} \\
\hline
MSCA-FFT (Ours) & \textbf{99.98} & \textbf{99.61} & \textbf{99.69} & \textbf{99.66} & \textbf{99.61} \\
MSCA-DCT~\cite{uddin2025msca} & 99.96 & 99.31 & 99.37 & 99.26 & 99.31 \\
Xception & 99.95 & 99.42 & 99.56 & 99.40 & 99.42 \\
ResNet50 & 94.39 & 86.25 & 85.37 & 89.10 & 86.46 \\
EfficientNetB4 & 86.02 & 76.66 & 76.51 & 85.53 & 78.19 \\
\hline
\end{tabular*}
\end{table}

Table~\ref{tab:method_comparison} compares MSCA-FFT with MSCA-DCT and several baseline CNN models on Celeb-DF (v2). The proposed MSCA-FFT model achieves 99.98\% AUC, 99.61\% accuracy, 99.69\% precision, 99.66\% recall, and 99.61\% F1-score. Overall, MSCA-FFT achieves the highest mean performance among the evaluated methods, including MSCA-DCT and the baseline CNN models. Compared with MSCA-DCT~\cite{uddin2025msca}, the proposed model obtains slightly higher mean values across all reported metrics. Although the improvement is modest, the consistent gain suggests that FFT-based spectral features can serve as an effective alternative to DCT-based frequency representations within the same spatial-frequency fusion framework.

To further evaluate robustness on a different benchmark, Table~\ref{tab:ffpp_results} reports the results on FaceForensics++. The proposed MSCA-FFT model achieves 
99.16\% AUC, 95.48\% accuracy, 96.92\% precision, 95.70\% recall, and 96.31\% F1-score. 
Compared with MSCA-DCT, the proposed model achieves higher mean performance across all reported metrics on this dataset. These results indicate that the FFT-based frequency representation generalizes beyond Celeb-DF (v2) and remains effective on a different deepfake benchmark.

\begin{table}[h]
\centering
\caption{Results on FaceForensics++ dataset.}
\label{tab:ffpp_results}
\footnotesize
\renewcommand{\arraystretch}{1.05}
\begin{tabular*}{0.98\columnwidth}{@{\extracolsep{\fill}}lccccc}
\hline
\textbf{Method} & \textbf{AUC} & \textbf{Acc.} & \textbf{Prec.} & \textbf{Rec.} & \textbf{F1} \\
\hline
MSCA-FFT (Ours) & \textbf{99.16} & \textbf{95.48} & \textbf{96.92} & \textbf{95.70} & \textbf{96.31} \\
MSCA-DCT~\cite{uddin2025msca} &  98.76                     & 94.46 & 95.38 & 95.64 & 95.51 \\
\hline
\end{tabular*}
\end{table}

\begin{figure*}[h!]
    \centering
    \begin{minipage}[t]{0.49\textwidth}
        \centering
        \includegraphics[width=\linewidth]{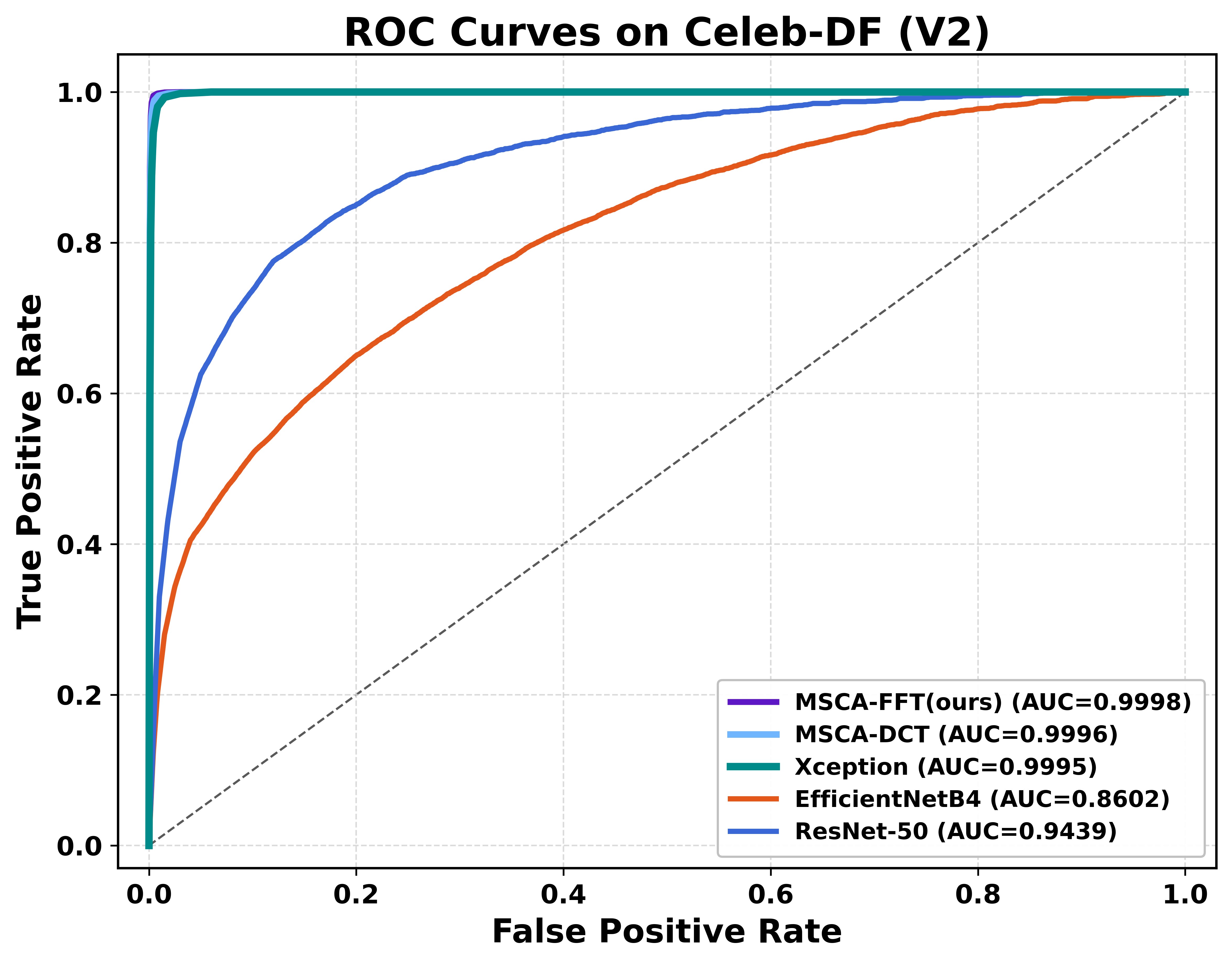}
        \vspace{2mm}
        \centerline{(a)}
    \end{minipage}
    \hfill
    \begin{minipage}[t]{0.49\textwidth}
        \centering
        \includegraphics[width=\linewidth]{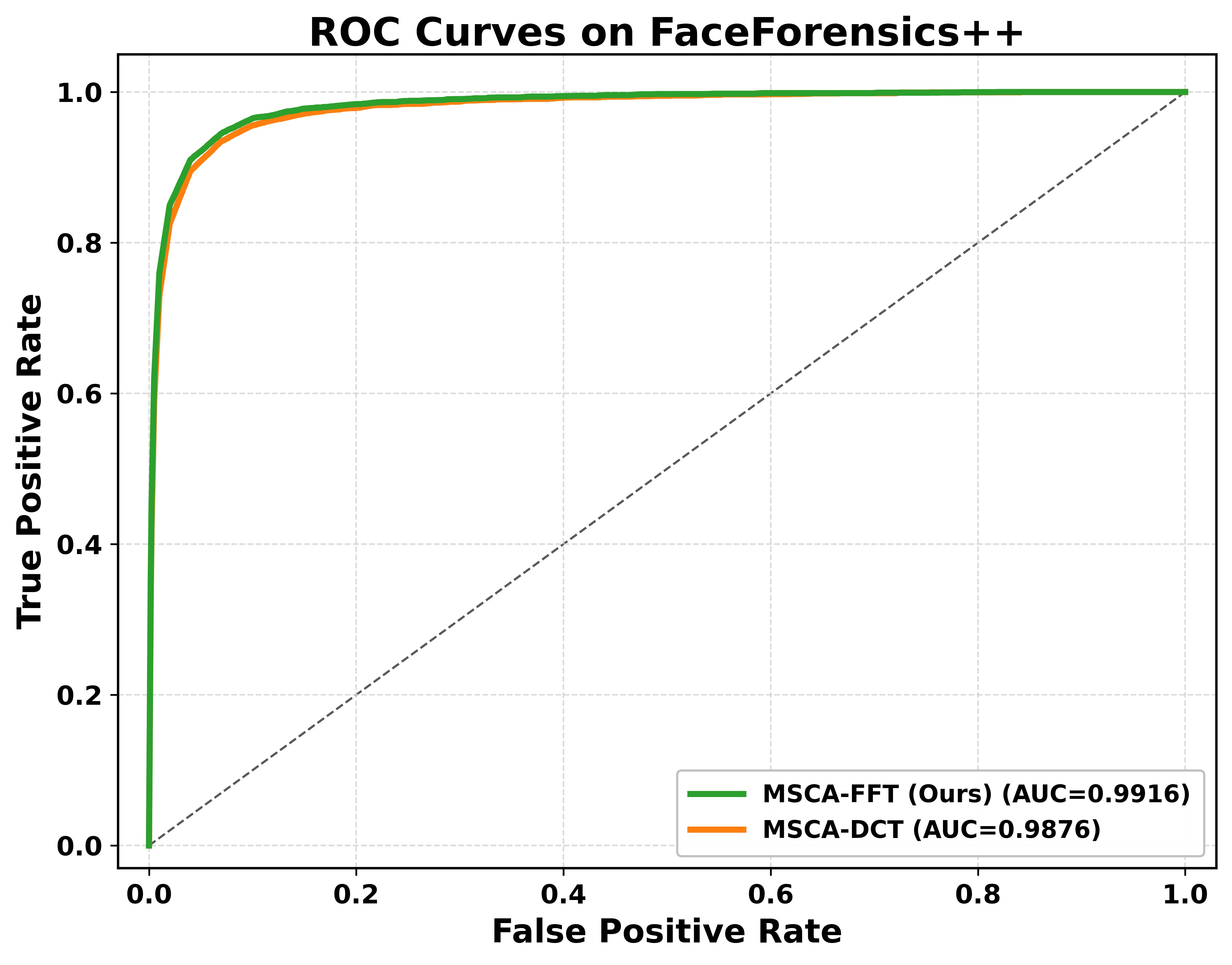}
        \vspace{2mm}
        \centerline{(b)}
    \end{minipage}
    \caption{ROC curves comparing MSCA-FFT with baseline and reference methods on two datasets: 
    (a) Celeb-DF (v2), corresponding to Table~\ref{tab:method_comparison}; 
    (b) FaceForensics++, corresponding to Table~\ref{tab:ffpp_results}.}
    \label{fig:roc}
\end{figure*}

The ROC curves in Fig.~\ref{fig:roc} provide a visual comparison of model performance on the two datasets. 
Fig.~\ref{fig:roc}(a) shows that both methods achieve near-saturated performance on Celeb-DF (v2), with ROC curves close to the upper-left corner and MSCA-FFT performing slightly better than MSCA-DCT. Fig.~\ref{fig:roc}(b) shows a more noticeable advantage for MSCA-FFT on FaceForensics++. Overall, the ROC curves are consistent with the quantitative results reported in Tables~\ref{tab:method_comparison} and~\ref{tab:ffpp_results}, confirming that MSCA-FFT maintains strong discriminative capability across both datasets.

\subsection{Ablation Study}

An ablation study was conducted to evaluate the contribution of the spatial and frequency components in the proposed MSCA-FFT framework. As summarized in Table~\ref{tab:ablation_study}, three variants were compared: the full model, the spatial branch only, and the frequency branch only. The spatial branch is implemented using the Xception backbone, with 80\% of its layers frozen during training. Thus, the spatial-branch-only variant removes the FFT-based frequency branch and the cross-attention fusion module. Similarly, the frequency-branch-only variant removes the Xception-based spatial branch and the cross-attention fusion module.

The frequency branch only achieves 67.78\% AUC, 62.22\% accuracy, 60.99\% precision, 73.67\% recall, and 66.26\% F1-score. This result indicates that frequency-domain information alone is not sufficient for robust deepfake detection. In contrast, the spatial branch only achieves 99.91\% AUC, 99.17\% accuracy, 99.30\% precision, 99.24\% recall, and 99.18\% F1-score, showing that spatial features provide strong discriminative cues for distinguishing real and manipulated faces. The full MSCA-FFT model achieves the best overall performance, with 99.98\% AUC, 99.61\% accuracy, 99.69\% precision, 99.66\% recall, and 99.61\% F1-score. 
Compared with the spatial-branch-only variant, the full model provides consistent improvements across all reported metrics. 
These results suggest that although the frequency branch is weaker as an independent classifier, it provides complementary spectral information when fused with Xception-based spatial features through the cross-attention fusion module.

\begin{table}[h]
\centering
\caption{Ablation study of the proposed MSCA-FFT framework on the Celeb-DF (v2) dataset. 
}
\label{tab:ablation_study}
\footnotesize
\begin{tabular*}{0.98\columnwidth}{@{\extracolsep{\fill}}lccccc}
\hline
\textbf{Variant} & \textbf{AUC} & \textbf{Acc.} & \textbf{Prec.} & \textbf{Rec.} & \textbf{F1} \\
\hline
Full model & \textbf{99.98} & \textbf{99.61} & \textbf{99.69} & \textbf{99.66} & \textbf{99.61} \\
Spatial branch only & 99.91 & 99.17 & 99.30 & 99.24 & 99.18 \\
Frequency branch only & 67.78 & 62.22 & 60.99 & 73.67 & 66.26 \\
\hline
\end{tabular*}
\end{table}

\subsection{Explainability Analysis}

\begin{figure*}[t]
  \centering
  \includegraphics[width= 1\textwidth]{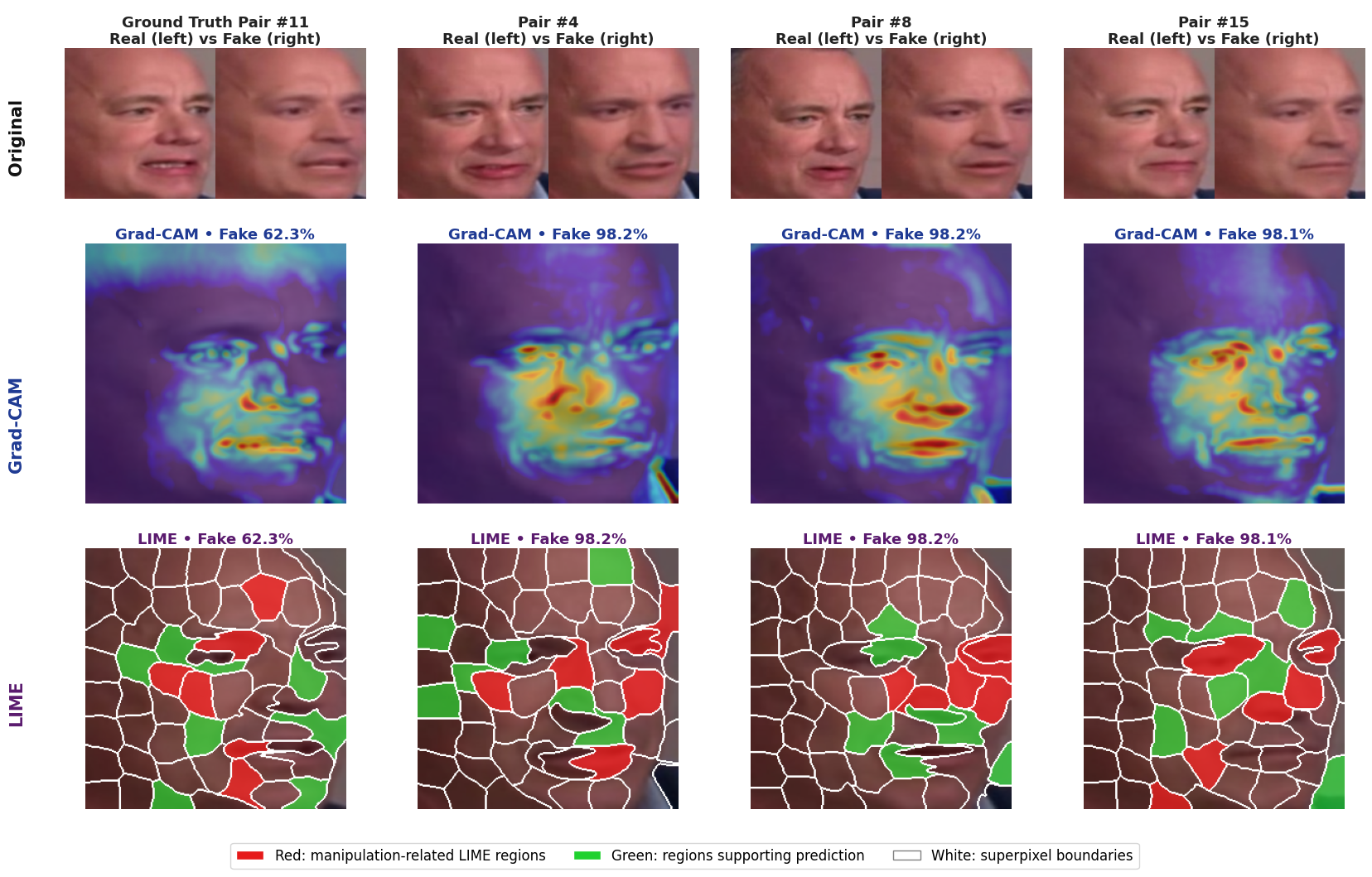}
  \caption{Grad-CAM and LIME visual explanations for representative real and fake image pairs. Grad-CAM highlights regions that strongly influence the classifier, while LIME identifies superpixels that support the fake prediction in green and regions that contradict it in red.}
  \label{fig:explainability}
\end{figure*}

Accurate classification alone is not sufficient for a reliable deepfake detector. The model should also base its decision on visual evidence that is
forensically meaningful. A common concern is that deep learning models may learn dataset-specific shortcuts, such as compression artifacts, capture
conditions, or other acquisition patterns, rather than the actual traces of facial manipulation. To assess this, we apply Grad-CAM and LIME to examine
whether the proposed model focuses on facial regions that are relevant to deepfake detection.

Grad-CAM~\cite{selvaraju2017gradcam} 
is applied to the final convolutional layer of the spatial branch to visualize image-domain regions that contribute to the fake-class prediction.
LIME~\cite{ribeiro2016lime} provides a complementary explanation by segmenting the face into superpixels using Simple Linear Iterative Clustering (SLIC). It then perturbs these regions and measures how each perturbation changes the model output. Superpixels that increase the
fake-class probability are interpreted as supporting evidence, while those that decrease it are interpreted as contradicting evidence.

As shown in Fig.~\ref{fig:explainability}, both explanation methods highlight facial areas that are important for manipulation analysis. The responses are mainly concentrated around the eye and eyebrow region, nose, mouth, and facial boundaries. These areas are significant because face manipulation methods often introduce blending inconsistencies, texture mismatch, and local geometric distortions in these regions. Pair~\#11 is classified as fake with a lower confidence of 62.3\%, and its explanations are more spatially distributed,
indicating a more difficult sample that lies closer to the decision boundary.
By contrast, Pairs~\#4, \#8, and~\#15 are classified with confidence scores above 98\% and show more localized Grad-CAM and LIME responses. This suggests that higher-confidence predictions are supported by clearer manipulation-sensitive evidence, while lower-confidence cases produce more diffuse explanations.

\section{Conclusion}\label{sec:conclusion}

This study presented MSCA-FFT, a spatial-frequency framework for image-level deepfake detection. The model combines a partially fine-tuned Xception spatial branch with a direct FFT-based frequency branch, followed by transformer-based feature modeling, cross-attention fusion, and MLP classification. Compared with the DCT-based MSCA method, the proposed frequency branch processes the log-scaled FFT magnitude spectrum directly and avoids inverse frequency-to-image reconstruction.

Experimental results on Celeb-DF (v2) and FaceForensics++ demonstrate that MSCA-FFT achieves strong detection performance.
On Celeb-DF (v2), the model achieved 99.98\% AUC, 99.61\% accuracy, 99.69\% precision, 99.66\% recall, and 99.61\% F1-score. On FaceForensics++, it achieved 99.16\% AUC, 95.48\% accuracy, 96.92\% precision, 95.70\% recall, and 96.31\% F1-score. 
The ablation study further shows that the spatial branch provides the strongest individual performance, while the frequency branch contributes complementary spectral cues when fused with spatial features.

The explainability results provide additional support for the proposed design. The FFT-based frequency analysis reveals spectral discrepancies around manipulation-sensitive facial regions, particularly near the eyes and mouth. This observation is consistent with the Grad-CAM and LIME results, which show that the model focuses on facial regions such as the eyes, nose, mouth, and boundaries when making real/fake predictions. Overall, these findings suggest that direct FFT-based frequency modeling, when combined with spatial features and cross-attention fusion, provides an effective and interpretable solution for deepfake face detection.

\bibliographystyle{IEEEtran}
\bibliography{references}

@inproceedings{chen2021crossvit,
  title     = {CrossViT: Cross-attention multi-scale vision transformer for image classification},
  author    = {Chen, Chun-Fu Richard and Fan, Quanfu and Panda, Rameswar},
  booktitle = {Proceedings of the IEEE/CVF International Conference on Computer Vision},
  pages     = {357--366},
  year      = {2021}
}

@article{dolhansky2019deepfake,
  title={The deepfake detection challenge (dfdc) preview dataset},
  author={Dolhansky, Brian and Howes, Russ and Pflaum, Ben and Baram, Nicole and Ferrer, Cristian Canton},
  journal={arXiv preprint arXiv:1910.08854},
  year={2019}
}

@inproceedings{chollet2017xception,
  title     = {Xception: Deep Learning with Depthwise Separable Convolutions},
  author    = {Chollet, Fran{\c{c}}ois},
  booktitle = {Proceedings of the IEEE/CVF Conference on Computer Vision and Pattern Recognition},
  pages     = {1251--1258},
  year      = {2017}
}

@inproceedings{afchar2018mesonet,
  title     = {MesoNet: A Compact Facial Video Forgery Detection Network},
  author    = {Afchar, Darius and Nozick, Vincent and Yamagishi, Junichi and Echizen, Isao},
  booktitle = {Proceedings of the IEEE International Workshop on Information Forensics and Security (WIFS)},
  pages     = {1--7},
  year      = {2018}
}

@inproceedings{yang2019headpose,
  title     = {Exposing Deep Fakes Using Inconsistent Head Poses},
  author    = {Yang, Xin and Li, Yuezun and Lyu, Siwei},
  booktitle = {Proceedings of the IEEE International Conference on Acoustics, Speech and Signal Processing (ICASSP)},
  pages     = {8261--8265},
  year      = {2019}
}

@inproceedings{frank2020leveraging,
  title={Leveraging frequency analysis for deep fake image recognition},
  author={Frank, Joel and Eisenhofer, Thorsten and Sch{\"o}nherr, Lea and Fischer, Asja and Kolossa, Dorothea and Holz, Thorsten},
  booktitle={International conference on machine learning},
  pages={3247--3258},
  year={2020},
  organization={PMLR}
}

@inproceedings{tan2024frequency,
  title={Frequency-aware deepfake detection: Improving generalizability through frequency space domain learning},
  author={Tan, Chuangchuang and Zhao, Yao and Wei, Shikui and Gu, Guanghua and Liu, Ping and Wei, Yunchao},
  booktitle={Proceedings of the AAAI Conference on Artificial Intelligence},
  volume={38},
  number={5},
  pages={5052--5060},
  year={2024}
}

@inproceedings{durall2020watch,
  title={Watch your up-convolution: Cnn based generative deep neural networks are failing to reproduce spectral distributions},
  author={Durall, Ricard and Keuper, Margret and Keuper, Janis},
  booktitle={Proceedings of the IEEE/CVF conference on computer vision and pattern recognition},
  pages={7890--7899},
  year={2020}
}

@inproceedings{rossler2019faceforensicspp,
  title     = {FaceForensics++: Learning to Detect Manipulated Facial Images},
  author    = {R{\"o}ssler, Andreas and Cozzolino, Davide and Verdoliva, Luisa and Riess, Christian and Thies, Justus and Nie{\ss}ner, Matthias},
  booktitle = {Proceedings of the IEEE/CVF International Conference on Computer Vision},
  pages     = {1--11},
  year      = {2019}
}

@article{xu2025few,
  title={Few-shot network intrusion detection method based on multi-domain fusion and cross-attention},
  author={Xu, Congyuan and Li, Donghui and Liu, Zihao and Yang, Jun and Shen, Qinfeng and Tong, Ningbing},
  journal={PloS one},
  volume={20},
  number={7},
  pages={e0327161},
  year={2025},
  publisher={Public Library of Science San Francisco, CA USA}
}

@article{Masood2023,
  author  = {Masood, Muhammad and Nawaz, Muhammad and Malik, Kamran M. and Javed, Ahsan and Irtaza, Ahsan and Malik, Hafiz},
  title   = {Deepfakes Generation and Detection: State-of-the-Art, Open Challenges, Countermeasures, and Way Forward},
  journal = {Applied Intelligence},
  year    = {2023},
  volume  = {53},
  number  = {4},
  pages   = {3974--4026}
}

@article{VaccariChadwick2020,
  author  = {Vaccari, Cristian and Chadwick, Andrew},
  title   = {Deepfakes and Disinformation: Exploring the Impact of Synthetic Political Video on Deception, Uncertainty, and Trust in News},
  journal = {Social Media + Society},
  year    = {2020},
  volume  = {6},
  number  = {1},
  doi     = {10.1177/2056305120903408}
}

@article{uddin2025msca,
  title={Spatial and frequency feature fusion using multi-scale cross attention for enhancing deepfake face detection: M. Uddin et al.},
  author={Uddin, Main and Fu, Zhangjie and Zhang, Xiang and Arnob, Abu Bakor Hayat},
  journal={Multimedia Systems},
  volume={31},
  number={4},
  pages={270},
  year={2025},
  publisher={Springer}
}

@article{shi2025customized,
  title={Customized transformer adapter with frequency masking for deepfake detection},
  author={Shi, Zenan and Chen, Haipeng and Jia, Yixin and Zhang, Dong and Lu, Wei and Yang, Xun},
  journal={IEEE Transactions on Information Forensics and Security},
  year={2025},
  publisher={IEEE}
}

@inproceedings{doloriel2024frequency,
  title={Frequency masking for universal deepfake detection},
  author={Doloriel, Chandler Timm and Cheung, Ngai-Man},
  booktitle={ICASSP 2024-2024 IEEE International Conference on Acoustics, Speech and Signal Processing (ICASSP)},
  pages={13466--13470},
  year={2024},
  organization={IEEE}
}

@inproceedings{yu2019attributing,
  title={Attributing fake images to gans: Learning and analyzing gan fingerprints},
  author={Yu, Ning and Davis, Larry S and Fritz, Mario},
  booktitle={Proceedings of the IEEE/CVF international conference on computer vision},
  pages={7556--7566},
  year={2019}
}

@article{guo2023constructing,
  title={Constructing new backbone networks via space-frequency interactive convolution for deepfake detection},
  author={Guo, Zhiqing and Jia, Zhenhong and Wang, Liejun and Wang, Dewang and Yang, Gaobo and Kasabov, Nikola},
  journal={IEEE Transactions on Information Forensics and Security},
  volume={19},
  pages={401--413},
  year={2023},
  publisher={IEEE}
}

@article{zhang2016mtcnn,
  title     = {Joint Face Detection and Alignment Using Multi-Task Cascaded Convolutional Networks},
  author    = {Zhang, Kaipeng and Zhang, Zhanpeng and Li, Zhifeng and Qiao, Yu},
  journal   = {IEEE Signal Processing Letters},
  volume    = {23},
  number    = {10},
  pages     = {1499--1503},
  year      = {2016}
}

@article{qiu2025d2fusion,
  title={D2Fusion: Dual-domain fusion with feature superposition for Deepfake detection},
  author={Qiu, Xueqi and Miao, Xingyu and Wan, Fan and Duan, Haoran and Shah, Tejal and Ojha, Varun and Long, Yang and Ranjan, Rajiv},
  journal={Information Fusion},
  volume={120},
  pages={103087},
  year={2025},
  publisher={Elsevier}
}

@inproceedings{Li2019Warping,
  author    = {Li, Yuezun and Lyu, Siwei},
  title     = {Exposing DeepFake Videos by Detecting Face Warping Artifacts},
  booktitle = {CVPR Workshops},
  year      = {2019}
}

@inproceedings{Li2020FaceXray,
  author    = {Li, Lingzhi and Bao, Jianmin and Zhang, Ting and Yang, Hao and Chen, Dong and Wen, Fang and Guo, Baining},
  title     = {Face X-Ray for More General Face Forgery Detection},
  booktitle = {CVPR},
  year      = {2020}
}

@inproceedings{li2020celeb,
  title={Celeb-df: A large-scale challenging dataset for deepfake forensics},
  author={Li, Yuezun and Yang, Xin and Sun, Pu and Qi, Honggang and Lyu, Siwei},
  booktitle={Proceedings of the IEEE/CVF conference on computer vision and pattern recognition},
  pages={3207--3216},
  year={2020}
}

@misc{celebdfv2kaggle,
  author= {Roy, Pranab},
  title= {Celeb-{DF} (V2) Image Dataset},
  howpublished = {\url{https://www.kaggle.com/datasets/pranabr0y/celebdf-v2image-dataset}},
  year= {2024},
  note= {Accessed: May 12, 2026}
}

@inproceedings{qian2020thinking,
  title={Thinking in frequency: Face forgery detection by mining frequency-aware clues},
  author={Qian, Yuyang and Yin, Guojun and Sheng, Lu and Chen, Zixuan and Shao, Jing},
  booktitle={European conference on computer vision},
  pages={86--103},
  year={2020},
  organization={Springer}
}

@inproceedings{selvaraju2017gradcam,
  title={{Grad-CAM}: Visual explanations from deep networks via gradient-based localization},
  author={Selvaraju, Ramprasaath R and Cogswell, Michael and Das, Abhishek and Vedantam, Ramakrishna and Parikh, Devi and Batra, Dhruv},
  booktitle={Proceedings of the IEEE International Conference on Computer Vision (ICCV)},
  pages={618--626},
  year={2017},
  doi={10.1109/ICCV.2017.74}
}

@inproceedings{ribeiro2016lime,
  title={``{W}hy should {I} trust you?'': Explaining the predictions of any classifier},
  author={Ribeiro, Marco Tulio and Singh, Sameer and Guestrin, Carlos},
  booktitle={Proceedings of the 22nd ACM SIGKDD International Conference on Knowledge Discovery and Data Mining},
  pages={1135--1144},
  year={2016},
  publisher={ACM},
  doi={10.1145/2939672.2939778}
}

\end{document}